\documentclass[10pt,conference,a4paper]{IEEEtran}
\usepackage{times,amsmath,epsfig}
\usepackage{epstopdf}

\usepackage{graphicx}
\usepackage{epsfig}
\usepackage{url}
\usepackage{multirow}
\usepackage{booktabs}
\usepackage{footnote}
\makesavenoteenv{tabular}

\title{Enhanced Local Binary Patterns for Automatic Face Recognition}

\author{
    {Pavel Kr\'al$^{1,2}$, Anton\'in Vrba$^{1}$, Ladislav Lenc$^{1,2}$}

\vspace{0.8cm}\\
\fontsize{10}{10}\selectfont%\itshape

{
  \begin{tabular}{cc}
  $^1$Dept. of Computer Science \& Engineering & $^2$New Technologies for the Information Society\\
  Faculty of Applied Sciences & Faculty of Applied Sciences\\
  University of West Bohemia & University of West Bohemia \\
  Plze\v{n}, Czech Republic & Plze\v{n}, Czech Republic\\
  \end{tabular}
}  

\,\\ 
\\
\fontsize{9}{9}\selectfont\ttfamily\upshape
\vspace{1.2mm}\\
\fontsize{10}{10}\selectfont\rmfamily\itshape
\,\\ 
\\

\fontsize{9}{9}\selectfont\ttfamily\upshape
\,
}

\begin{document}
%\ninept

\maketitle

\vspace{-3cm}

% INCLUDES COPYRIGHT NOTICE: one of three copyright notice should be included. Uncomment the appropriate line below, according to the authors affiliation:
\begin{figure}[b]
\parbox{\hsize}{\em
%information about the event:
%IEEE VCIP'14, Dec. 7 - Dec. 10, 2014, Valletta, Malta.

%copyright notice: one of three copyright notices below should be included. Uncomment the appropriate line, according to the authors affiliation:
%000-0-0000-0000-0/00/\$31.00 \ \copyright 2014 IEEE.
%U.S. Government work not protected by U.S. copyright.
%???-?-????-????-?/10/\$??.?? \copyright 2014 Crown.
}\end{figure}

\begin{abstract}
This paper presents a novel automatic face recognition approach based on local binary patterns. This descriptor considers a local neighbourhood of a pixel to compute the feature vector values. This method is not very robust to handle image noise, variances and different illumination conditions. We address these issues by proposing a novel descriptor which considers more pixels and different neighbourhoods to compute the feature vector values. The proposed method is evaluated on two benchmark corpora, namely UFI and FERET face datasets. We experimentally show that our approach outperforms state-of-the-art methods and is efficient particularly in the real conditions where the above mentioned issues are obvious. We further show that the proposed method handles well one training sample issue and is also robust to the image resolution.
\\[1\baselineskip]
\end{abstract}

% NOTE keywords are not used for conference papers so do not populate them

\section{Introduction}
%\vskip -3pt
\label{sec:introduction}
Automatic face recognition (AFR) consists in person identification from digital images using a computer.
This field has been intensively studied during the past a few decades and its importance is constantly growing particularly due to the nowadays security issues.

It has been proved that local binary patterns (LBP) are an efficient image descriptor for several tasks in computer vision field %~\cite{ojala1996comparative}
including automatic face recognition~\cite{Ahonen04}. %todo--add Ladas LBP paper
It considers a very small local neighbourhood of a pixel to compute the feature vector values.
The individual values are then computed using the differences between intensity values of the central and surrounding pixels.

%pav-why we did not used LBP with learning
In this paper, we propose a novel image descriptor called {\em Enhanced local binary patterns (E-LPB)}.
This method improves the original LBP operator by considering larger central area and larger neighbourhood to compute the feature vector values.
These properties keep more information about the image structure and can compensate some noise, image variance issues and the differences between train / test images.
This method of computation of the LBP operator considering more points has, to the best of our knowledge, never been done before and it is thus the main contribution of this paper.

The proposed method is evaluated on two standard corpora, UFI~\cite{LencUfi2015} and FERET~\cite{Phillips98} face datasets.
UFI dataset is chosen to show the results in real conditions where the images are noisy, vary in the pose and are illuminated differently.
FERET corpus is used in order to show the results of one training sample issue.
In this case, we have only one image for training.
Therefore it is not possible to improve the results by training step as presented for instance in~\cite{lei2014learning} and we focus thus rather on the descriptor itself.

%The rest of the paper is organized as follows.
%The following section describes the most important methods based on LBP.
%Section~\ref{sec:met} details the proposed approach.
%Section~\ref{sec:exp} first describes the corpora used for evaluation and then presents the results of experiments realized on this data.
%The last section discusses the results and proposes some future research directions.

\section{{Related Work}}
%\vskip -3pt
\label{sec:relwork}
Methods based on local binary patterns generally use LBP histograms computed in rectangular regions~\cite{Ahonen04}.
The concatenated histograms create face representation vectors which are then compared using a distance metric.
%s as for instance histogram intersection or Chi square distance~\cite{zhang2003evaluation}.
Uniform local binary patterns are an interesting LBP extension~\cite{ojala1996comparative} which reduces the histogram size to 59.
%The number of uniform patterns is 58 and all non-uniform patterns belong to the 59th bin. 
Ojala et al.~\cite{ojala2002multiresolution} further use a~circular neighbourhood created by a~number of points $P$ placed on a circle with a diameter $R$. This LBP variant is denoted as LBP$_{P,R}$.

Li et al.~\cite{li2012face} propose dynamic threshold local binary pattern (DTLBP).
They use the mean value of the neighbouring pixels and also the maximum contrast between the neighbouring points to compute the feature vector. 
Another LBP extension are local ternary patterns (LTP)~\cite{tan2010enhanced} which uses three states to capture the differences between the central pixel and the neighbouring ones.
%The authors claim that both DTLBP and LTP are less sensitive to the noise than the original LBP method.

Local derivative patterns (LDP) are proposed in~\cite{zhang2010local}.
The difference from the original LBP is that it uses the features of higher order.
%It thus can represent more information than the original approach.
Davarzani et al.~\cite{davarzani2014robust} propose a~weighted and adaptive LBP-based texture descriptor.
This approach successfully handles some issues of the previously proposed LBP-based approaches such as invariance to scaling, rotation, viewpoint variations and non-rigid deformations.

Elongated binary patterns~\cite{liao2007face} are another variant of the LBP using an elliptical instead of circular neighbourhood.
The main advantage of this modification is that it retains better structural information in the images.
Jin et al.~\cite{jin2004face} propose improved local binary patterns (ILBP).
This method compares the intensities of neighbourhood pixels against the local mean pixel intensity (instead of the intensity of the central pixel).
%This reduces the effect of noise.
%Generalized Local Binary Patterns for Texture Classification ~\cite{li2011generalized}
%https://uwaterloo.ca/vision-image-processing-lab/sites/ca.vision-image-processing-lab/files/uploads/files/lily_bmvc2011_3.pdf

%Extended local binary patterns for texture classification ~\cite{liu2012extended}
%http://www.sciencedirect.com/science/article/pii/S0262885612000066
Another interesting LBP adaptation proposed by Li et al. is extended local binary patterns~\cite{liu2012extended}.
This method introduces two different and complementary feature types (pixel intensities and differences).
%Experimental results demonstrate significant improvements over the classical LBP approach in texture classification task.

%\cite{barkan2013fast}

%ICAE2016
The previously described methods were oriented to the modification of the LBP operator itself, however creation of the feature vector and recognition procedure remain usually similar.
Both tasks are significantly improved by Lenc and Kral~\cite{lenc2015local} by automatic identification of the important facial points using Gabor wavelets and k-means clustering algorithm.
%The feature vectors are then created in such positions and the features are compared individually. % (instead of creation of one large vector).
%It was experimentally shown that this method is very efficient on several standard face corpora.
Lei et al. ~\cite{lei2014learning} further propose a learning step to improve the results of the LBP operator when more gallery images available.
%Discriminant image filters learning and optimal soft neighbourhood sampling are used for learning task.

%For additional information about the LBP based methods, please refer to the survey~\cite{huang2011local}. %,nanni2012survey}.

\section{Enhanced  Local Binary Patterns for Face Recognition}
%\vskip -3pt
\label{sec:met}

\subsection{Local Binary Patterns}
%\vskip -3pt
\label{sec:lbp}
The original LBP~\cite{ojala2002multiresolution} operator uses a~3$\times$3 square neighbourhood centred at a given pixel.
The algorithm assigns either 0 or 1 value to the 8 neighbouring pixels by Equation~\ref{eq:lbp_1}.
%This algorithm is illustrated in Figure~\ref{fig:lbp_computation}.

%\begin{figure}[!htb]
%  \centering
%  {\epsfig{file=lbp_computation.pdf,width=6.5cm,angle=0}}
%   \caption{An example of the feature computing by the original LBP operator}
%  \label{fig:lbp_computation}
%\end{figure}

\begin{equation}
\label{eq:lbp_1}
N_i =
\left\{
	\begin{array}{ll}
		0 & \mbox{if } {g_i < g_c} \\
		1 & \mbox{if } {g_i \geq g_c}
	\end{array}
\right.
\end{equation} 
\linebreak
where $N_i$ is the binary value assigned to the neighbouring pixel $i\in \{1,..,8\}$,
$g_i$ denotes the gray-level value of the neighbouring pixel $i$ and $g_c$ is the gray-level value of the central pixel. 
The resulting values are then concatenated into an 8 bit number. 
Its decimal representation is used to create the feature vector.

\subsection{Enhanced Local Binary Patterns (E-LBP)}
%\vskip -3pt
\label{sec:elbp}
We extend the original LBP operator by computing the feature values from point-sets instead of the isolated points.
We also consider different sizes of the neighbourhood of the central area.
This concept can handle several LBP issues:
%LBP issues
\begin{itemize}
\item LBP has small spatial support, therefore it cannot properly detect large-scale textural structures;
\item It loses local textural information, since only the signs of differences of neighbouring pixels are used;
\item It is sensitive to noise, because the slightest fluctuation above or below the value of the central pixel is treated as equivalent to a major contrast between the central pixel and its surroundings.
\end{itemize}

\begin{figure}[!h]
  \centering
  {\epsfig{file=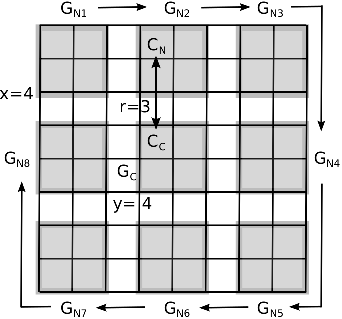,width=4.5cm,angle=0}}
   \caption{Scheme of E-LBP$_{4,4,3}$ operator (i.e. x=4, y=4, r=3)}
  \label{fig:elbp_computation}
\end{figure}

The proposed algorithm is depicted in Figure~\ref{fig:elbp_computation} and simultaneously described next.
Let ${G_{N_i}}$ be a~set of neighbouring pixel intensities with its central pixel $C_{N_i}$ (the closest left/top pixel is used as the central one in the case of the neighbourhoods of the even size).
Let $G_C$ be a~set of central pixel intensities with its central pixel $C_C$
and let $r$ be a~distance between the central pixels $C_{N_i}$ and $C_C$.
We calculate the representative values for these sets as average values of the pixel intensities belonging to these sets: $g'_{i} = mean (G_{N_i})$, $i\in \{1,..,8\}$ and $g'_C= mean (G_C)$.

The feature vector is then created in a similar way as in the case of the original LBP operator using $g'_i$ and $g'_C$ values instead of $g_i$ and $g_c$, respectively (see Section~\ref{sec:lbp}).

Note that it is possible to consider several point-set topologies of different sizes to capture different texture information, however in this paper we use only the square shapes of the sizes $2 \times 2$, i.e. $4$ points and $3 \times 3$ points, i.e. $9$ points.

The proposed operator is further denoted as E-LBP$_{x,y,r}$, where $x \in \{4,9\}$ represents the neighbouring pixel-set topology,
$y \in \{4,9\}$ is the central pixel-set topology and $r$ is the distance between the central pixels $C_N$ and $C_C$, which is hereafter called {\em E-LBP range}.

The source codes of this algorithm are freely available for research purposes at~\url{http://home.zcu.cz/~pkral/sw/}.

\subsection{Face Modelling and Recognition}
\label{sec:face_modelling}
%\vskip -3pt
We compute LBP values in all points of the face image.
The image is then divided into a~set of square cells lying on a~regular grid. 
Feature vectors are computed for each cell as a~histogram of the E-LBP values.
Every cell is then represented by one feature vector of the size 256.
%\subsection{Face Comparison}
As many other LBP based face recognition methods,
we concatenate the feature histograms into one feature vector to create the face model.
%According to our previous experiments~\cite{lenc2015local}
We use a histogram intersection distance  with 1-NN classifier for the face recognition. % defined as: 

%pav-todo
%http://users.monash.edu.au/~dengs/resource/papers/icnnsp03.pdf -- HI definition

%\begin{equation}
%\label{eq:intersect}
%HI(f,r)=1 - \sum_i{\min({f_i, r_i})}
%\end{equation}

%where $f$ and $r$ represent the faces (represented by feature vectors) to compare and $i$ is the number of histogram bins.
%We use this form of HI to be able to interpret it as a distance where 0 value means the identical histograms.

\section{Evaluation}
%\vskip -3pt
\label{sec:exp}

\subsection{Experimental Set-up and Corpora}
%\vskip -3pt
We used OpenCV\footnote{http://opencv.org/} toolkit for implementation of our models to realize the following experiments.
The face databases used for evaluation of our approach are briefly described next.

\subsubsection{UFI Dataset}
Unconstrained facial images (UFI) dataset~\cite{LencUfi2015} contains face images of 605 persons extracted from real photographs and is mainly dedicated for face recognition in real conditions. % (noisy images with varying head pose, different illumination conditions, etc.).
%The gallery set contains about 7 images per person while one face image is designed for testing.
In the following experiments, we use {\it Cropped images} partition. 
Figure~\ref{fig:example_cropped} (left) shows two images of one individual from this partition with recognition results of our method.

\subsubsection{FERET Dataset}
\begin{figure}[!h]
  \centering
 {
\epsfig{file=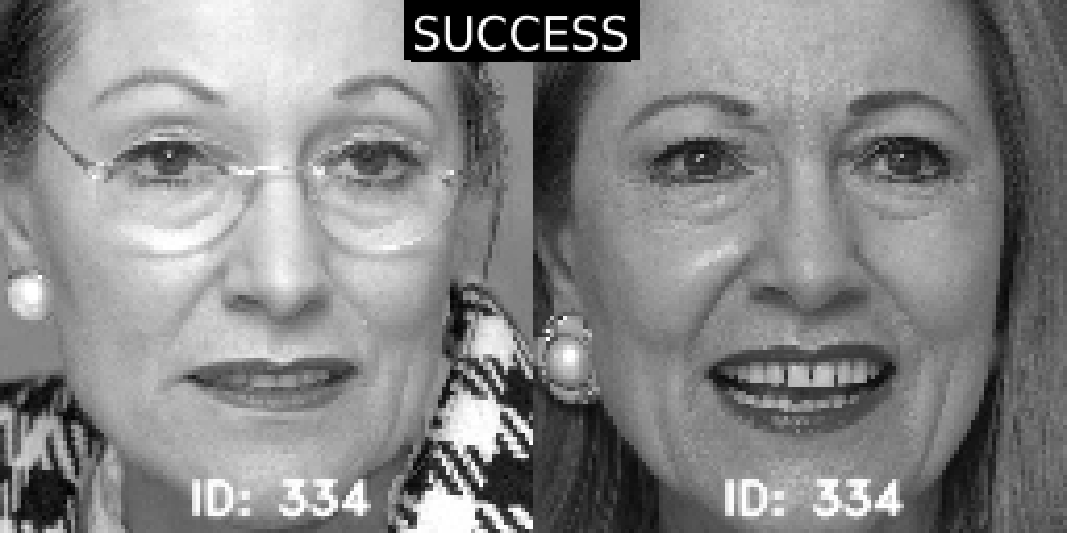,width=4.1cm,height=2.0cm}
\epsfig{file=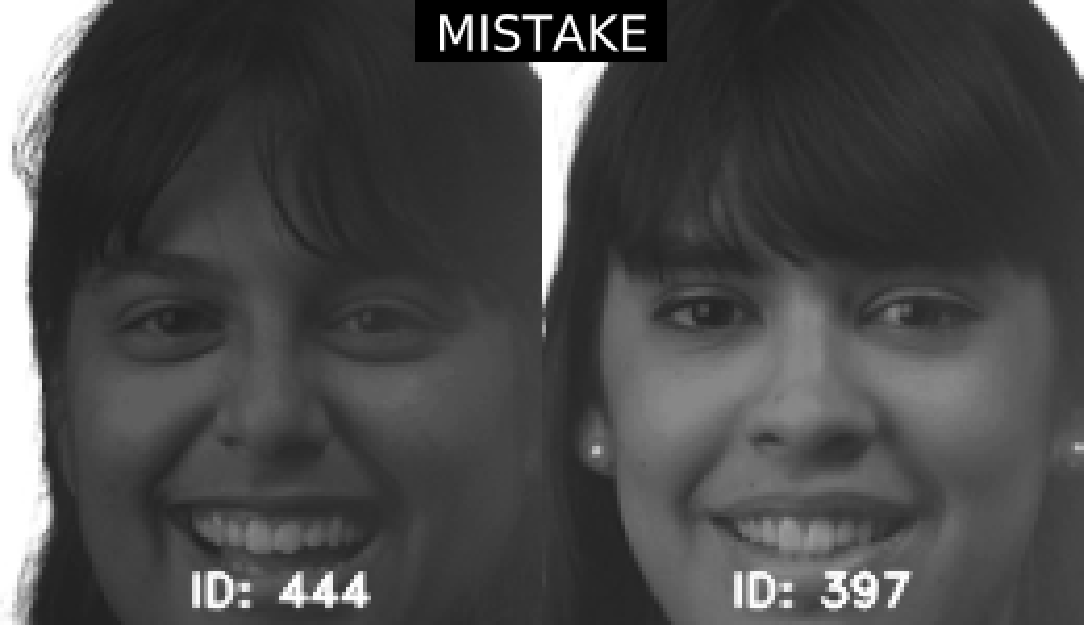,width=3.8cm,height=2.0cm}
%UFI
%  \epsfig{file=cropped_1.eps,width=1.9cm,angle=0}
%  \epsfig{file=cropped_2.eps,width=1.9cm,angle=0}
%  \epsfig{file=cropped_1.eps,width=2cm,height=2.5cm,angle=0}
%  \epsfig{file=cropped_2.eps,width=2cm,height=2.5cm,angle=0}
%FERET
%\epsfig{file=feret_example_1.eps,width=1.9cm,height=2.5cm,angle=0}
%  \epsfig{file=feret_example_2.eps,width=1.9cm,height=2.5cm,angle=0}
%
  }
   \caption{Example images of one person from the UFI (left, correct recognition) and FERET (right, recognition error) datasets}
  \label{fig:example_cropped}
\end{figure}

%\vskip -3pt
\label{sec:feret}
FERET dataset~\cite{Phillips98} contains 14,051 images of 1,199 individuals. 
%The images are divided into the several probe sets according to the face pose.
We use {\it fa} set for training while {\it fb} set for testing of the proposed method which represents 1195 of different individuals to recognize.
Note that only one image per person/set is available therefore we address the one training sample problem.
For the following experiments, the faces are cropped according to the eye positions and resized to $130\times150$ pixels.
Figure~\ref{fig:example_cropped} (right) shows two example images of one person from the FERET database with recognition results obtained by the proposed approach.

\subsection{Optimal Cell Size of the Proposed Approach}
%\vskip -3pt
The cell size (see Section~\ref{sec:face_modelling}) is one important parameter of the whole approach.
This value should be set correctly to obtain a good recognition accuracy.
However, it does not influence the E-LBP operator itself and it should depend mainly on the image resolution. 
We thus set this value experimentally using original LBP operator.

\begin{figure}[!htb]
  \centering
  {\epsfig{file=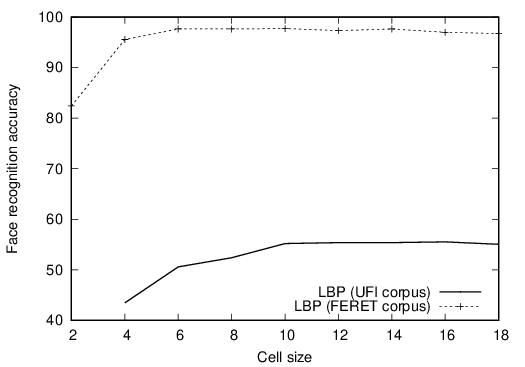,width=9.0cm,angle=0}}
   \caption{Face recognition accuracy of the LBP on UFI and FERET corpora depending on the cell size}
  \label{fig:xp1}
% \vspace{-0.3cm}
\end{figure}

The results of this experiment are depicted in Figure~\ref{fig:xp1} for UFI and FERET corpora.
This figure shows that the recognition accuracies are increasing and from the value of 10 they remain almost constant for both corpora.
Therefore, we chose this value for the following experiments.

%\vskip -3pt
\subsection{Optimal Range of the Proposed Operator}
\begin{figure}[!htb]
  \centering
   {\epsfig{file=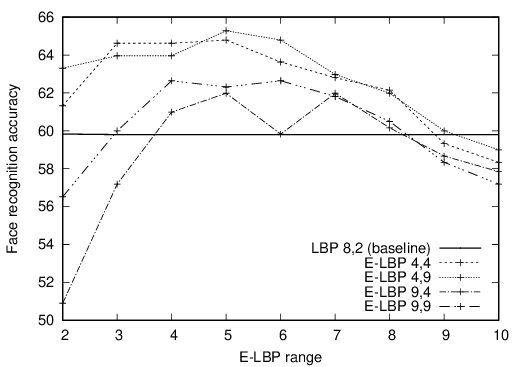,width=9.0cm,angle=0}}
 %  \vspace{-5pt}
   \caption{Face recognition accuracy of the proposed method on UFI dataset depending on the E-LBP range}
  \label{fig:xp2}
%\vspace{-3pt}
\end{figure}

\begin{figure}[!htb]
  \centering
   {\epsfig{file=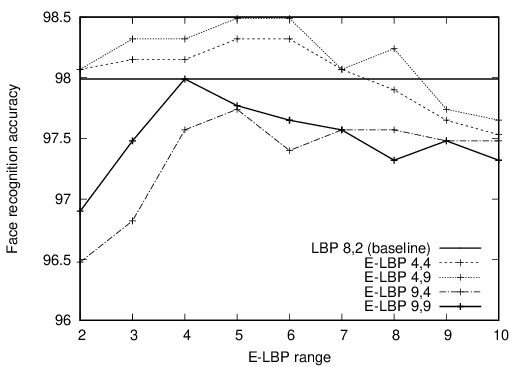,width=9.0cm,angle=0}}
 %  \vspace{-5pt}
   \caption{Face recognition accuracy of the proposed method on FERET dataset depending on the E-LBP range}
  \label{fig:xp2f}
% \vspace{-3pt}
\end{figure}

%\vskip -3pt
E-LBP range (see Section~\ref{sec:elbp}) is another important parameter of the proposed method.
It defines the distance between the individual point-sets to compute the feature vector values and
it also influences significantly the recognition results.
Therefore, we determine its optimal value for both corpora in the second experiment (see Figure~\ref{fig:xp2} for UFI and Figure~\ref{fig:xp2f} for FERET dataset).
We can thus summarize:
%\vspace{-5pt}
\begin{itemize}
\item The optimal E-LBP range is 5 for both corpora;
\item The best topology is E-LBP$_{4,9}$ for both corpora;
\item The results of E-LBP$_{4,4}$ are almost similar to E-LBP$_{4,9}$; 
\item Proposed E-LBP operator significantly outperforms the baseline LBP in these two cases on both corpora;
\item The behaviour of this operator on both corpora is consistent (similar progress).
\end{itemize}

We conclude that the proposed E-LBP operator is very robust and we also assume that it should perform well on other corpora using these settings.

\vskip -15pt 
\subsection{Image Resolution Evaluation}
\begin{figure}[!htb]
  \centering
   {\epsfig{file=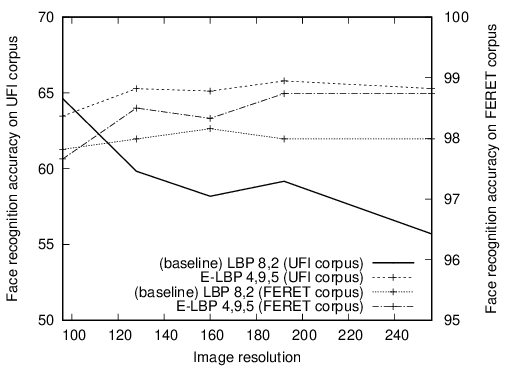,width=8.8cm,angle=0}}
 %  \vspace{-5pt}
   \caption{Face recognition accuracy of the proposed method compared to the baseline LBP$_{8,2}$ operator depending on the image resolution on UFI and FERET corpora}
  \label{fig:xpRes}
 %\vspace{-3pt}
\end{figure}

Another important requirement is the robustness to the different image resolution.
It is beneficial to keep the high recognition score also when image resolution is changing.
Therefore, we report in Figure~\ref{fig:xpRes} the dependence of the recognition accuracy on the image resolution.
The image resolution varies from 96$\times$96 to 256$\times$256 and we use both corpora for this experiment.

This figure shows that the proposed E-LBP approach is robust against the image resolution on both corpora.
The recognition accuracy is higher than the baseline LBP$_{8,2}$ operator, except for the $96\times96$ case.
We can thus conclude that the proposed operator is not suitable for images in very small resolution.

\vskip -15pt 
\subsection{Final Results}
\vskip -2pt
Table~\ref{tab:final_res1} compares the performance of the proposed method against several other state-of-the-art algorithms. % (see Table~\ref{tab:final_res1}).
It demonstrates that the proposed approach is efficient particularly in the real conditions 
(i.e. UFI dataset), where it outperforms the standard LBP by 10\% and the previous best method by 2\% in absolute value.
%This method also achieves the best recognition rate on FERET dataset, however the improvement is only marginal, because the results are already close to 100\%.
This method also achieves competitive recognition rate on FERET dataset (one training sample issue).
%Although the score is not the highest one, it is comparable with the best method, because the difference is not statistically significant.

\begin{table}[htb]
%\label{tab:final_res1}
\centering
%\small{
\footnotesize{
\begin{tabular}{l|cc}
\toprule
  & \multicolumn{2}{c}{Recognition rate [\%]} \\
Approach & UFI & FERET \\
\midrule
SRC (Wagner et al.~\cite{Wagner12})  & - & 95.20 \\
LBP (Ahonen et al.~\cite{Ahonen04}) & 55.04 & 93.89 \\
LBP$_{8,2}$ & 59.83 & 97.99 \\
uniform LBP$_{8,2}$ & 53.39 & 97.66 \\
LDP (Zhang et al.~\cite{zhang2010local}) & 50.25 & 97.4 \\
FS-LBP (Lenc et al.~\cite{lenc2015local}) & 63.31 & {\bf 98.91} \\
\midrule
E-LBP$_{4,9,5}$ \scriptsize{(proposed)} & {\bf 65.28} & { 98.5} \\ %--todo-check res on FERET
\bottomrule
\end{tabular}
}
\vskip -1pt
\caption{Final results of the proposed approach on the UFI and FERET databases against several state-of-the-art methods}
\label{tab:final_res1}
\end{table}

%\vskip -25pt
\section{Conclusions} % and Future Work}
%\vskip -2pt
\label{sec:conclusion} %pav-TODO -- uncomment footnote for CR!!
This paper introduced a novel face recognition approach based on LBP. 
We proposed an original image descriptor which considers more pixels and different neighbourhoods to compute the feature vector.
We evaluated this method on the standard UFI and FERET face datasets.
The source codes are freely available for research purposes at~{\it url\_hidden\_for\_review}. %~\url{http://home.zcu.cz/~pkral/sw/}.

We experimentally showed that our approach outperforms a~number of other state-of-the-art methods (LBP$_{8,2}$ included) and its capabilities are particularly evident in the real conditions when images can be noisy, vary in the pose and are illuminated differently.
We also demonstrated that the proposed approach is robust to the image resolution.
This was demonstrated on the UFI dataset, where we obtained recognition accuracy 65.28\%,
which represents the increase by 2\% over the other best method.

%The first perspective consists in evaluation of different point-set topologies (see Section~\ref{sec:elbp}) to compute the feature vector.
%Then, we will modify the matching method as suggested in~\cite{lenc2015local}. 
%We also would like to evaluate the proposed method on some other corpora and also on other tasks (e.g. texture classification or object recognition) to demonstrate its robustness and applicability to other domains.

%\vfill\eject

%\section{Acknowledgment}
%This work has been supported by the project LO1506 of the Czech Ministry of Education, Youth and Sports.

\bibliographystyle{./IEEEtran}

\bibliography{paper}

\end{document}